\pdfoutput=1
\documentclass[11pt]{article}
\usepackage{acl2023}

\usepackage{times}
\usepackage{latexsym}

\usepackage[T1]{fontenc}
\usepackage[utf8]{inputenc}

\usepackage{microtype}
\usepackage{booktabs}
\usepackage{graphicx}
\usepackage{enumitem}
\usepackage{algorithm}
\usepackage{algpseudocode}
\usepackage{amsfonts}
\usepackage{amsmath}
\usepackage{multirow}
\usepackage{caption}
\usepackage{subcaption}
\usepackage{bbm}

\title{Few Shot Rationale Generation using Self-Training with Dual Teachers}

\author{
	Aditya Srikanth Veerubhotla$^{1}$\thanks{\hspace{4pt} Work done during an internship at Amazon}
	\And Lahari Poddar $^2$ 
	\AND  Jun Yin$^2$
	\And  György Szarvas $^2$  \\ 	\\
	$^1$Language Technologies Institute, Carnegie Mellon University\\
	{\tt adityasv@cs.cmu.edu} \\
	$^2$Amazon\\
	{\tt \{poddarl, jnyin, szarvasg, sharanye\}@amazon.com} \\
	\And Sharanya Eswaran $^2$ 
	}

\begin{document}
\maketitle

\begin{abstract}
%Prior works on self-rationalizing models, which generate free-text rationales for their predicted labels rely on large, manually annotated corpora. We study the self-rationalization problem in the context of few-shot labels and develop a self-training approach leveraging both labeled and unlabeled data.  We propose a novel dual-teacher learning framework that learns two specialized teacher models for task prediction and rationalization using self-training. The knowledge from these teachers is distilled into a multi-tasking student model that can jointly generate the task label and rationale. 
%We conduct extensive experiments on three public datasets to evaluate the proposed method and demonstrate its effectiveness in modeling task labels and generating faithful rationales. 
%Our results suggest that self-training is a promising direction to train generative NLP models with limited labeled data while leveraging pseudo labels on larger unlabeled data.

Self-rationalizing models that also generate a free-text explanation for their predicted labels are an important tool to build trustworthy AI applications. Since generating explanations for annotated labels is a laborious and costly process, recent models rely on large pretrained language models (PLMs) as their backbone and few-shot learning.
In this work we explore a self-training approach leveraging both labeled and unlabeled data to further improve few-shot models, under the assumption that neither human written rationales nor annotated task labels are available at scale. 
We introduce a novel dual-teacher learning framework, which learns two specialized teacher models for task prediction and rationalization using self-training and distills their knowledge into a multi-tasking student model that can jointly generate the task label and rationale. 
Furthermore, we formulate a new loss function, Masked Label Regularization (MLR) which promotes explanations to be strongly conditioned on predicted labels.
Evaluation on three public datasets demonstrate that the proposed methods are effective in modeling task labels and generating faithful rationales. 
%Our results suggest that self-training is a promising direction to train generative NLP models with limited labeled data while leveraging pseudo labels on larger unlabeled data.
\end{abstract}

\section{Introduction}
Interpretable NLP has emerged to learn models which explain their predictions through either extractive \cite{deyoung2020eraser} or natural language explanations \cite{esnli, Narang2020WT5TT, Wiegreffe2020MeasuringAB}. Due to higher expressivity of free text, generative self-rationalizing models have gained much research interest. %Sequence to sequence models \cite{t5} have shown to be capable of jointly producing task label and explanation as a single sequence for different tasks. 
However, the early works assume a fully supervised setup and require a large annotated dataset \cite{Narang2020WT5TT}. 
Collecting large scale, manual annotations for task labels and corresponding explanations is challenging and expensive. On the other hand, a much larger unlabeled corpora is often available,  making semi-supervised approaches like few-shot learning \cite{brown2020languagemodels} and self-training \cite{he2019revisiting}  attractive solutions.
In the context of self-rationalizing models, \cite{marasovic-etal-2022-shot} explore few-shot learning, while \citep{zelikman2022star} seek to improve a supervised labeler by augmenting it with rationale generation.
% in a self-training loop.
In this work we start from a few-shot setup, assuming only a handful of examples available with their labels and hand-written rationale. 
We leverage a large unlabeled dataset and self-training techniques to improve over the simple few-shot model.

%We propose a novel Dual Teacher learning framework to learn a self-rationalizing model from two teacher models. 
%We hypothesize that learning a joint model for generating both task label and explanation is much harder compared to learning conditionals, especially in limited supervision settings. The proposed approach therefore first learns two models through self-training in a cascading manner. Predictor model for predicting task labels, and a Rationalizer model which generates an explanation given an input and the task labels from the Predictor. 
We hypothesize that using only a few examples, learning to generate meaningful explanations \emph{jointly} with predicting the labels themselves, is a particularly challenging objective and self-training can suffer from a weak initial model.  
To address this, we propose a novel Dual Teacher learning approach to learn a self-rationalizing model from the two teacher models in a cascading manner. 
At first, a Predictor model is learned for predicting task labels, and then a Rationalizer model is learned to generate an explanation conditioned on an input and the task labels predicted by the Predictor model. We iteratively improve both models via self-training. 
%and utilize them to learn a strong Joint model capable to perform both tasks. 
In contrast to learning the Joint model directly, the Rationalizer model allows for much richer representation learning by moving the label information from decoder to the encoder part, and utilizing the encoder's self-attention mechanism to extract input-label correlations. 
A stronger few-shot model for rationale generation provides higher quality pseudo labels, consequently making self-training more effective.

Although the two conditional models (Predictor and Rationalizer) might be better performing, a single self-rationalizing model is still desirable for practical applications, due to its ease-of-maintenance and parameter efficiency for faster inference. 
We apply principles from knowledge distillation \cite{Hinton2015DistillingTK, kim2016sequence} on the two conditional models to learn a joint model that generates task label and explanation as a single sequence. 
The teacher models are used for generating pseudo labels on the entire unlabeled dataset. The initial few-shot labeled data and the pseudo labeled dataset are finally combined to train the joint model.

Faithfulness of explanations is an imperative property for practical applications of interpretability analysis. A model generated explanation is considered faithful if it accurately explains the decision making of the model \cite{alvarez2018towards, Wiegreffe2020MeasuringAB}. 
Similar to prior study \cite{jacovi2020towards}, we also observe that a free text explanation generated by models might sound \textit{plausible}, without satisfying the \textit{faithfulness} criteria of explaining the predicted task label. 
This motivates us to design a masking based regularization function, Masked Label Regularizer (MLR), to encourage the model to condition on the task label while generating an explanation. 
%We construct an entropy based constraint that forces the Rationalizer model to be maximally uncertain in generating an explanation in absence of label tokens. 
MLR is an entropy based constraint that forces the Rationalizer model to be maximally uncertain in generating an explanation in absence of label tokens and is used to ensure that the Rationalizer model preserves faithfulness through the self-training iterations.
To summarize, our contributions are:

\begin{itemize}[leftmargin=*]
\item Proposing to utilize self-training for learning self-rationalizing models with free-text explanations, demonstrating that it provides significant performance boost compared to few-shot learning.
\item Proposing a novel Dual Teacher framework, where two teacher models are trained with self-training in a cascading manner for learning two tasks, and a multi-task joint student model is learned through distillation from the teachers.
\item Extensively studying the faithfulness property of free-text explanations, and designing an entropy based regularization to encourage label-explanation conditioning.
\item Experiments on three public benchmark datasets and demonstrating the effectiveness of our proposed model in improving both task accuracy and explanation quality.
\end{itemize}

\section{Related Work}
%Self-rationalizing models can be broadly categorized into models producing extractive or free-text rationales. 
Prior works on generating free text rationales have explored joint models \cite{Narang2020WT5TT, marasovic-etal-2022-shot} as well as several variants of pipeline models \cite{Wiegreffe2020MeasuringAB, pte}.  We also use sequence to sequence models  \cite{t5} as our backbone models. 
While most of the self-rationalizing literature assumes fully supervised setups, STaR \cite{zelikman2022star} explores an alternate bootstrapping setup where limited rationales are available, but the task labels are present for the whole dataset. We consider the generic and more restrictive setting where only limited annotations are available for both task label and rationale.

For limited labeled data scenario, many NLP applications have started reporting success with self-training \cite{mehta2022improving, yu-etal-2022-actune, he2019revisiting, bhat-etal-2021-self}. Inspired from these works, we employ self-training to the self-rationalization problem.
We introduce a new training framework with two conditional models and using them as teachers in a further distillation step to train the joint model. 
Besides the popular use for model compression, Knowledge Distillation has also shown superior performance when using the same model architecture and size for both the student and teacher models \cite{furlanello2018born}, and distilling from multiple teachers \cite{yuan2021reinforced, liu2020adaptive}.
%, distilling a single student model from multiple teachers specialized in multiple tasks \cite{clark2019bam, ghiasi2021multi}. 
Recently, a work \cite{ghiasi2021multi} in computer vision domain has explored using pseudo-labels from multiple teachers to train a joint student model. However, they have multiple specialized teachers trained independently through full supervision, in contrast to the cascading nature of our dual teacher self-training setup.

Evaluating the quality of free-text rationales is significantly challenging and several works have proposed metrics to evaluate the explanations around fluency and their faithfulness properties \cite{hase2020evaluating, hase2020leakage, marasovic-etal-2022-shot}. A recent work \cite{wang2022pinto} also tries to imbue faithfulness through a regularizing coefficient. However, they apply the regularizer to perturb the rationale while generating task label. In contrast we use a label masking regularizer to enforce the Rationalizer model to generate an explanation which is faithful to the label.

\begin{figure}[tbp]
	\centering 
	\includegraphics[width=0.98\linewidth]{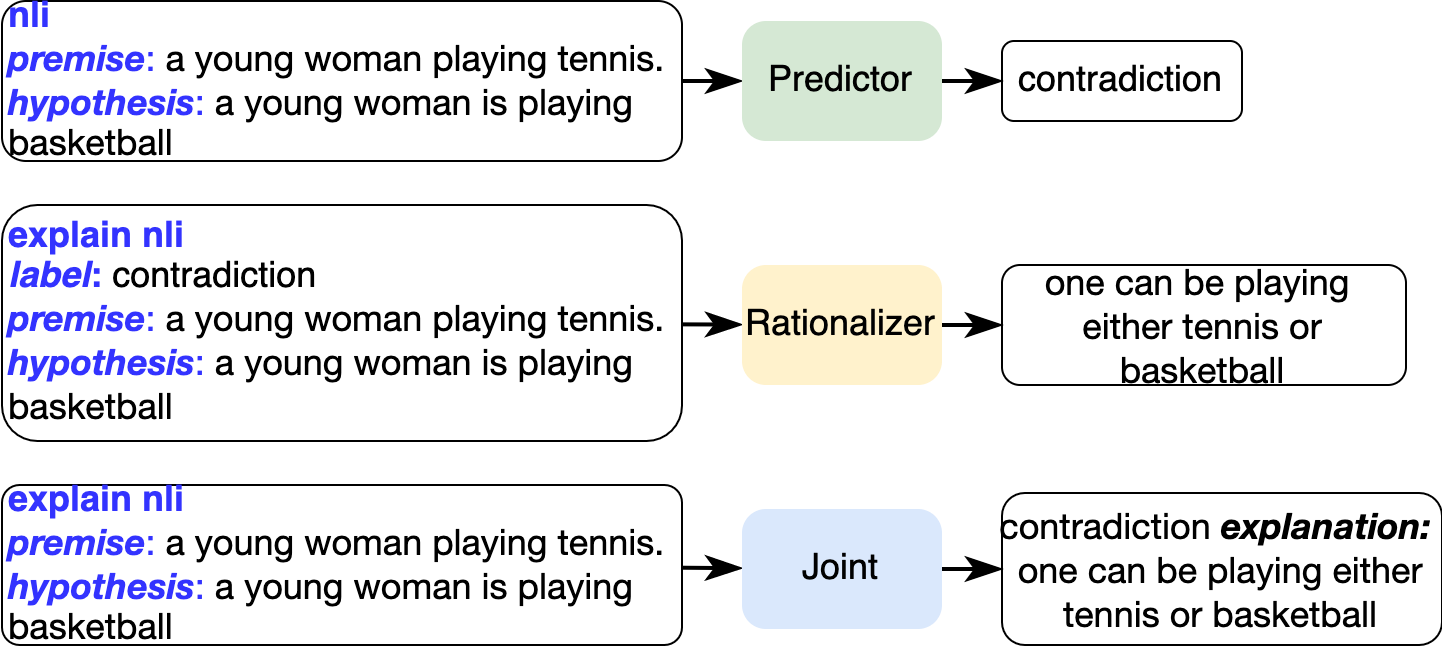}
	\caption{Input and output formats for Predictor, Rationalizer and Joint models.}
	\label{fig:io_format}
\end{figure}

\section{Background}
We first provide some necessary background on Self-Rationalizing models and a theoretical outline of Self Training based learning.

\noindent \textbf{Self-Rationalization}: 
A Self-Rationalization model tries to learn the joint distribution of output($O$) and explanation($E$), given an input($I$), i.e. $P(O,E|I)$. A common approach is modeling it as a sequence-to-sequence problem and generating the task prediction and the rationale jointly \cite{Narang2020WT5TT}.
%We follow the formulation used in WT5 \cite{Narang2020WT5TT}, where given an input sequence $i = \{x_1, x_2, \cdots x_T\}$ of $T$ tokens, the model produces an output sequence $o = \{y_1, y_2, \cdots y_K\}$ in an auto-regressive manner. 
Input-output format for a self rationalizing joint model is illustrated in Figure \ref{fig:io_format}. The input consists of a task prompt, (e.g. \texttt{explain nli}), and in output sequence the task label is generated first (e.g. \texttt{contradiction}), followed by a separator token (\texttt{explanation:}), and then the free text explanation. During inference, greedy decoding is used to generate the sequence until an EOS token is produced.

% \begin{equation}
% P(O,E|I) = P(O|I)P(E|I,O)
% \end{equation}

% Factorization of the joint distribution in the above way means the output is predicted first, and the explanation is generated conditioned on both the input and the output label. We use a 

%
%\begin{algorithm}[htbp]
%\small
%\caption{SelfTraining}\label{alg:self_training_definition}
%\begin{algorithmic}
%\Require $D_l$ \Comment{Labeled dataset} 
%\Require $D_u$ \Comment{Unlabeled dataset}
%\Require $D_{val}$ \Comment{Validation dataset}
%\Require $\theta^T, \theta^S$ \Comment{Initial parameters}
%
%\While{ not converged } 
%    \State $\theta^T \gets Train(D_l, D_{val}, \theta^T)$
%    \State $D_{pl} \gets \{(x_i, p_{\theta^T}(x_i))\}_{i=1}^{|D_u|}$
%    \State $\theta^S \gets Train(D_{pl}, D_{val}, \theta^S)$ 
%    \State $\theta^T \gets \theta^S$
%\EndWhile
%\end{algorithmic}
%\end{algorithm}

\noindent \textbf{Self-Training} is a type of Semi-Supervised Learning based method, which assumes access to a small labeled dataset ($D_l$) and a large, unlabeled in-domain dataset ($D_u$).  
The algorithm progresses iteratively in four steps. First, a teacher model is trained on the labeled dataset ($D_l$), to obtain  $\theta^T$. The trained teacher is then used to infer \textit{pseudo-labels} on $D_u$, generating the \textit{pseudo-labeled} dataset $D_{pl}$. A student model is then trained on $D_{pl}$ to obtain the $\theta^S$. In the next iteration the teacher model is updated with the learned parameters from the student and the process repeats until a convergence criterion is met. 
%Algorithm \ref{alg:self_training_definition} describes the self-training procedure.

\section{Dual Teacher for Self-Rationalization}

\begin{figure*}[htbp]
\centering \includegraphics[width=0.85\linewidth]{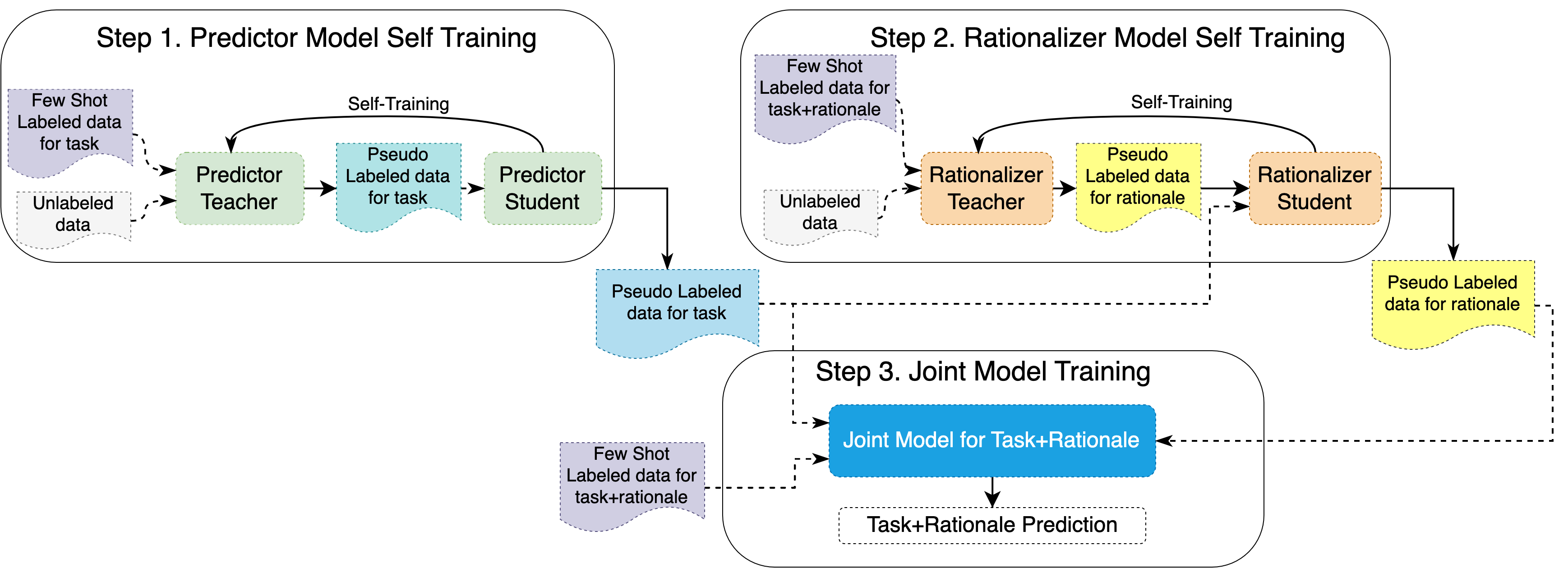}
\caption{Dual Teacher Training Framework. Predictor and Rationale models are trained in their own Self-training loop. Pseudo labels generated from the trained predictor and rationale model are used for training the Joint model.}
\label{fig:dual_teacher}
\end{figure*}

We combine the strengths of self-training and knowledge distillation to train a self-rationalizing joint model from dual teachers. 
%First, we train two teacher models, Predictor and Rationalizer using self-training. Then a student joint model is trained using soft-outputs of the teachers as pseudo labels. 
Following sections describe the components, their losses and the learning procedures in more detail. Input-output formats of the models are shown in Figure \ref{fig:io_format}, and the overall framework is illustrated in Figure \ref{fig:dual_teacher}.

\subsection{Problem Setup}
We tackle the self-rationalization problem with few-shot labels. We consider access to a small labeled set, $ D_l = \{(i_j, o_j, e_j)\}_{j=1}^N $, where $i_j$ is the input, $o_j$ is the task output, and $e_j$ is the natural language explanation. We also leverage a much larger unlabeled dataset denoted by $D_u = \{i_j\}_{j=1}^M$, where $M \gg N$. In the unlabeled dataset only the input text is available and no annotation is provided for either task label or rationale.

%The likelihood loss for generating a sequence is computed as
%\begin{equation}
%\mathcal{L} = \prod_{j=1}^K p(y_j|x_1, \cdots, x_T, y_1, y_2, \cdots y_{j-1})
%\end{equation}
%\noindent where $y_j$ is the predicted token at the $j^{th}$ position in the output sequence, $y_1, y_2, \cdots y_{j-1}$ are the tokens preceding it, $x_1, x_2, \cdots, x_T$ are input tokens, and $K$ is the length of the output sequence.

To keep all models identical, we model all distributions in a sequence to sequence manner using T5 \cite{t5}.
The teacher model in self-training is trained on few shot ground truth output sequences and the trained teacher is then used for generating output sequences for the unlabeled dataset. These sequences are considered as pseudo labels to train the student model.
We re-weight the loss of each example with confidence of the teacher model. 
%In other words, up-weight examples where teacher is confident and down-weight predictions with poor confidence. 
This limits error propagation through self-training iterations due to the noisy nature of pseudo labels. We use likelihood of the generated sequence as confidence estimates. Following \cite{bhat-etal-2021-self} we normalize the weights in a batch.
%, such that the weights across the batch sum to $1$. 

\subsection{Splitting the Joint into Conditionals}

In order to make the learning task easier, we break down the joint probability of modeling task and rationale, into its conditionals.
\begin{equation}
    \underbrace{P(O,E|I)}_\text{Joint} = \underbrace{P(O|I)}_\text{Predictor} \times \underbrace{P(E|I,O)}_\text{Rationalizer}
\end{equation}
This allows us to build two separate models in a cascading manner: (1) Predictor Model for predicting task label, i.e. $P(O|I)$, and (2) Rationalizer Model for rationalizing the task label for an input, i.e. $P(E|I,O)$.
Prior works \cite{pte} have shown that factorization of this distribution to predicting the output first (Prediction) and generating an explanation for the prediction (Rationalization) has obtained better performance than alternate factorizations.

We hypothesize that with limited labeled examples, learning a joint distribution for <task label+rationale> sequence would be much harder than focusing on learning to predict only the task label. 
More importantly, for rationale generation we move the task label from output sequence (in the joint model) to input sequence (in Rationalizer model). This allows the encoder to capture much richer interactions between task label and the input through its self-attention network, compared to only the decoder in joint model.
The stronger initial few-shot models for predictor and rationalizer would be  further boosted through self-training in generating higher quality pseudo labels.

\subsection{Predictor Teacher}
In the first step of our framework, we train a Predictor model with self-training. The Predictor is trained to model the probability of the task output given the input, i.e. $P(O|I)$.  The task output is decomposed into subwords, and the model is trained to minimize the negative log likelihood of the output token sequence:
\begin{equation}
\mathcal{L}_{pred}(\theta) = \mathbb{E} _{(i,o) \sim \mathcal{D}} \left[ -\log P_{\theta}(o|i) \right]
\end{equation}
The predictor model is trained within its own self training loop, utilizing the few shot ground truth task labels and unlabeled inputs. After self-training has converged, we store the predictor  and use it for generating pseudo task labels on all unlabeled data.
\begin{equation}
    D_{pl} = \{(i, p_{\theta_{pred}}(o|i))\}_{i \in D_u}
\end{equation}
These pseudo labels are then used for training the Rationalizer model and the Joint model.
 
\subsection{Rationalizer Teacher}
In the second stage we train a Rationalizer model that can  generate natural language explanations given an input and the predicted task output, modeling the conditional distribution $P(E|I,O)$. 
%We conduct self-training on another T5 model, where we append the task output to the input. 
\begin{equation}
\mathcal{L}_{rat\_gen}(\theta) = \mathbb{E} _{(i,o, e) \sim \mathcal{D}} \left[ -\log P_{\theta}(e|i,o) \right]
\end{equation}
For training the teacher model we use the few-shot ground truth labeled dataset for task label and rationale. For generating rationale pseudo-labels on the unlabeled set, we use the task pseudo labels generated by the predictor model as input. The generated rationale pseudo labels are then used to train a student rationalizer model in self-training loop until convergence.

\subsubsection*{Faithfulness of Explanations}
For a Rationalizer model to generate a \textit{faithful} explanation, we want the explanation to be strongly conditioned on the label. 
The rationalizer should not be able to generate an explanation solely based on the input, but must take into consideration the label for which it is rationalizing. 
We introduce a regularizing constraint in our rationalizer model to explicitly encode this property.

%This defeats the faithfulness of the explanations generated by the Rationalizer, leading to poor pseudo-labels for the Joint student. 

\subsubsection*{Masked Label Regularization}

We design an entropy based regularization which tells the model to be maximally uncertain in generating the explanation in absence of a task label. We achieve this by replacing the task output with mask tokens and maximizing the per-token entropy of the explanation sequence.
\begin{equation}
\mathcal{L}_{MLR}(\theta) = \mathbb{E}_{(i,e) \sim \mathcal{D}} \left[- H_{\theta}[e|i] \right]
\end{equation}
\noindent where $H_{\theta}[e|i]$ refers to the entropy of producing an explanation from input directly. 

There could be alternate ways of encoding the constraint of label-explanation association. We experimented with one such variant where the ground truth explanation would be generated with a high entropy in case of a wrong label. We observed similar empirical results in our experiments for this alternative. However, it is strictly less general - since it becomes limited to only categorical problems, and also is computationally more expensive due the necessity of computing entropy for multiple wrong labels. Therefore, we use the simpler and generic form of masking the label tokens.

The overall loss of the Rationalizer is a weighted summation of the sequence generation loss and the regularization loss:
\begin{equation}
\mathcal{L}_{rat} = \mathcal{L}_{rat\_gen}(\theta) + \lambda_{MLR}\mathcal{L}_{MLR}(\theta)    
\end{equation}
\noindent $\lambda_{MLR}$ is empirically set to $1e^{-4}$ in our experiments for all datasets.

\begin{algorithm}[tbp]
\small
\caption{Dual Teacher Training Algorithm}\label{alg:dual_teacher_self_training_definition}
\begin{algorithmic}
\Require $D_l = \{(i_i, o_i, e_i)\}_{i=1}^N$
\Require $D_u = \{i_j\}_{j=1}^M$
\Require $D_{val} = \{(i_k, o_k, e_k)\}_{i=1}^K$
\State Initialize $\theta_{pred}$, $\theta_{rat}$, $\theta_{joint}$ randomly\\
\\
/* Train Predictor model */
\State $\theta_{pred}^{*} \gets SelfTraining(\mathcal{D}_l, \mathcal{D}_u, D_{val}, \theta_{pred})$
\State $\mathcal{D}_{pred} \gets \{(i_j, \hat{o}_j)\}_{j=1}^M, \hat{o}_j \sim p_{\theta_{pred}^{*}}(.|I)$ \\
\\
/* Train Rationalizer model */
\State $\theta_{rat}^{*} \gets SelfTraining(\mathcal{D}_{l}, \mathcal{D}_{pred}, D_{val}, \theta_{rat})$
\State $\mathcal{D}_{pl} \gets \{(i_j, \hat{o}_j, \hat{e}_j)\}_{j=1}^M,$\\
$ \hat{o}_j \sim p_{\theta _{pred}^{*}}(.|I), \hat{e}_j \sim p_{\theta _{rat}^{*}}(.|I,O)$  \\
\\
/* Train Joint model */
\State $D_{final} \gets D_{pl} \cup D_l$
\State $\theta_{joint}^{*} \gets Train(D_{final}, D_{val}, \theta_{joint})$
\end{algorithmic}
\end{algorithm}

\subsection{Learning from Multiple Teachers: Distilling a Joint from the Conditionals}
\label{sec:dual_teacher_joint}

Knowledge Distillation is an effective learning paradigm to train a lighter student model with rich supervision signals from  better performing teacher model(s).
To alleviate the limitations of limited labeled data for learning a good self-rationalization model, we leverage the unlabeled data and collect task and rationale pseudo-labels sequentially from trained Predictor and Rationalizer teacher models. 
The final pseudo-labeled dataset is then combined with the few-shot labeled data and a joint model is trained on this set. 
This allows the knowledge from both the Predictor and Rationalizer models to be distilled into the student Joint model through pseudo labels and the teachers' confidence weights.

The joint model is trained to maximize the likelihood of a concatenated sequence of task output and explanation, as illustrated in Figure \ref{fig:io_format}.  
The detailed training algorithm is described in Algorithm \ref{alg:dual_teacher_self_training_definition}.

%We first explore training a joint model that generates task output and explanation in a self-training framework. 
%An "\textit{explanation:}" prompt is used to distinguish the output tokens and explanation tokens. 

\noindent \textbf{Loss Re-weighting}: Similar to most sequence-to-sequence models, in WT5 \cite{Narang2020WT5TT}, all output tokens in the generated sequence have uniform weights in the loss. However, in the joint task setup, the number of tokens from task label is substantially smaller than those in the explanation. 
%This would imply that with uniform weighting, the explanation component would have more contribution to the loss than the task label. 
To balance this, we re-weight the token-level losses between the output and the explanation. For a tuple $(i_j, o_j, e_j)$, the loss is computed as:
\begin{multline*}
\mathcal{L}  = \lambda\sum_{y_m \in o_j}-\log p_{\theta}(y_m|i_j, y_1, \cdots y_{m-1}) \\ + (1-\lambda)\sum_{y_n \in e_j}-\log p_{\theta}(y_n|i_j, y_1, \cdots y_{n-1}) 
\end{multline*}
\noindent where $\lambda \in [0.5, 1)$ is a weight coefficient.

\section{Results and Discussion}
We evaluate on public datasets for three different tasks. Table \ref{tab:data_stat} shows statistics of the datasets.

\noindent \textbf{e-SNLI} \cite{esnli}  extends the popular SNLI dataset \cite{bowman-etal-2015-large} by adding human-annotated explanations to the NLI 
labels. The task requires generation of a task label which describes the relationship between a premise and a hypothesis as entailment/contradiction/neutral, and a free text explanation for the prediction. 

\noindent \textbf{ComVE} \cite{comve}  aims to evaluate if a model can distinguish between sensible and nonsensical statements based on common knowledge. We combine the data from SubTask A (Validation) and SubTask C (Generation) for our experiments. 

\noindent \textbf{ECQA} \cite{ecqa} augments the Commonsense QA dataset \cite{commonsenseqa} with free-text explanations that support the the correct answer choice and refute the incorrect ones. We utilize the explanations for the correct output (Positive Property) as the explanation.

For few-shot settings we sample $100$ examples per class for each dataset. The self-training setup leverages the few-shot labeled dataset($D_l$) and the rest of the training set as unlabeled dataset($D_u$).

\begin{table}[tbp]
	\centering
	\resizebox{0.9\linewidth}{!}{
		\begin{tabular}{l|ccc}
			\hline
			& e-SNLI & ComVE & ECQA \\ \hline
			\# classes & 3 & 2 & 5 \\
			total train size & 549,367 & 10,000 & 7,598 \\
			few shot dataset size & 300 & 200 & 500 \\
			validation size & 9,842 & 1,000 & 1,090 \\
			test size & 9,824 & 1,000 & 2,194 \\
			Avg. tokens in output & 2.0 & 2.0 & 1.9 \\
			Avg. tokens in explanation & 16.8 & 26.0 & 14.5 \\ \hline
		\end{tabular}
	}
	\caption{Dataset Statistics. Token-level statistics were generated using the T5-base tokenizer.}
	\label{tab:data_stat}
\end{table}

\begin{table*}[tbp]
	\centering
	\resizebox{0.85\textwidth}{!}{
		\begin{tabular}{lcccccccc}
			\toprule
			Model & \multicolumn{2}{c}{e-SNLI} & \multicolumn{2}{c}{ComVE} & \multicolumn{2}{c}{ECQA} & \multicolumn{2}{c}{Average} \\
			/Metric & Acc & BLEU & Acc & BLEU & Acc & BLEU & Acc & BLEU\\
			\midrule
			%\textit{Random} & 33.33 & - & 50.0 & - & 20 & - & 34.44 & - \\
			%\textit{WT5}  & 90.9 & 32.4 & - & - & - & - & - & - \\
			\textit{Fully Supervised} \cite{Narang2020WT5TT} & 90.44 & 33.76 & 86.2 & 14.53 & 53.6 & 16.25 & 76.75 & 21.5 \\ 
			\textit{Few-Shot} & 82.57 & 24.21 & 73.77 & 12.74 & 34.29 & 9.77 & 63.54 & 15.57 \\
			\midrule
			\textbf{Self-Training techniques} &&&&&&&&\\
			\textit{Vanilla} & 83.35 & 25.18 & 78.83 & 10.44 & 41.8 & 9.7 & 67.99 & 15.11 \\
			\textit{Confidence Weighted} & 83.41 & 24.54 & 79.23 & 11.04 & 41.75 & 9.85 & 68.13 & 15.14 \\
			\textit{Dual Teacher} & \textbf{83.95} & \textbf{30.17} & \textbf{79.61} & \textbf{14.83} & \textbf{44.26} & \textbf{17.12} & \textbf{69.27} & \textbf{20.71} \\
			\bottomrule
	\end{tabular}}
	\caption{Results on different baselines and other self-training techniques on three datasets, measured using Accuracy for label prediction, and BLEU for explanation}
	\label{tab:main_results}
\end{table*}

\subsection{Implementation Details}
We use the base variant of T5 \cite{t5} as backbone model for the Predictor, Rationalizer and Joint models. 
%Owing to computational resources, we used the base variant of T5. 
%Figure \ref{fig:io_format} describes the formats for the input and output for the three models.
Following \cite{Narang2020WT5TT}, we also measure task performance using accuracy, and rationalization using SacreBLEU \cite{sacrebleu}. Label smoothing was set to 0.1 and early stopping with a patience of 5 was used for model selection. 
The few-shot examples were sampled randomly by stratifying across classes. 
We trained on 4 NVIDIA v100-16GB GPUs with a batch size of $8$ and $16$ for $D_l$ and $D_{pl}$, respectively.
The token re-weighting coefficient $\lambda$ is set to $0.8$ for eSNLI and ComVE, and $0.9$ for ECQA via grid search based on validation scores and average length of the explanations in the dataset. 
All results are reported after averaging $3$ runs.
%The input and output formats for the different datasets are described in \ref{sec:dataset_io_formats}.
\subsection{Main Results}
\label{subsec:main_results}

In Table \ref{tab:main_results} we compare the various training paradigms, namely, fully supervised, few-shot training, and self-training on all three datasets. 
For self-training we explore two setups - one without pseudo label re-weighting on a Joint model, which we call Vanilla Joint. Confidence-weighted Joint performs self-training on a Joint model where the pseudo labels are weighted by the confidence of the teacher model. The Dual Teacher refers to the proposed Joint model in Section \ref{sec:dual_teacher_joint} that is trained with distillation from two teachers.

\noindent \textbf{Few-shot vs Fully Supervised results:} Metrics from the fully-supervised setup provide an upper bound on the scores achievable when trained on complete dataset of labels and rationales. 
%As expected there exists a large gap of about $10-19\%$ in accuracy and $2-9$ points in BLEU between the few-shot and fully supervised setup across datasets.
%The performance in e-SNLI being withing 10\% of the fully supervised model, to a 12\% difference in performance in the case of ComVE and a large difference of 19\% in the case of ECQA. 
%Similarly, there is a large variance in the rationalization performance, with the smallest gap of 2 BLEU in the case of ComVE and the largest in e-SNLI, with a gap of 9 BLEU. 
Aggregated across datasets, the few-shot performance of the model is around $13\%$ behind the fully supervised model, and around $5$ BLEU lower in rationalization performance. 

\noindent \textbf{Self-Training helps boosting few-shot results.} Our experiments show that self-training is a promising direction in bridging the performance gap, improving accuracy and BLEU across all the tasks over the  few-shot counterparts.
We observe that re-weighting the pseudo-labels with the confidence of the teacher models, provides small improvements in the overall performance and is in alignment with previous findings \cite{bhat-etal-2021-self}. 

\noindent \textbf{Stronger Results with Dual Teacher Self-Training Framework.} Finally, we observe a further improvement by our proposed method of performing self-training on the Predictor and Rationalizer models, and subsequently distilling the knowledge to a joint student model through pseudo labels. The improvement in aggregate scores shows that the accuracy is within $8\%$ of a fully supervised model, and $5\%$ higher than the few-shot baseline. 
The improvements from the proposed model are most prominent for the Rationale generation task - the BLEU scores are improved by a large margin compared to learning both tasks jointly in a self-training setup. Impressively, the dual-teacher approach achieves an aggregated result of $20.71$ BLEU which is close to the aggregate performance of the \textit{Fully Supervised} model ($21.5$ BLEU). We even obtained higher performance (BLEU score) than the supervised model on the two smaller datasets, ComVE and ECQA.
%This is in line with our hypothesis of higher sample complexity for obtaining a strong Joint model. 

% Things to compare:
%   1. few-shot and fully supervised
%   2. vanilla vs conf
%   3. predictor vs vanilla/conf
%   4. rationalizer vs vanilla/conf
%   5. rationalizer-faithfulness vs rationalizer
%   8. joint-student vs joint-student + faithfulness

% [Lahari]
% \textbf{Questions}
% \begin{itemize}
%     \item Are the numbers in main table significant?
%     \item Is the faithfulness regularizer applied to the student as well? or only the rationalizer teacher \\
%         \textbf{[Aditya]} Only teacher. Need to explain why this did not work in the joint model setup
%     \item `Joint + Dual Teacher -- pipeline + faithfulness` has a lower BLEU score than `Joint + Dual Teacher -- pipeline`. Is this significant or some glitch in training?  \\
%         \textbf{[Aditya]} Yes, the predictor's labels were getting deleted when the data was stored and subsequently read. Fixed it now, getting results
%     \item Define the consistency metric for faithfulness evaluation of the student. \\
%         \textbf{[Aditya]} Two part definition: \\
%             1. Dependence on the label: num(examples with all different explanations per label)/num(total examples) \\
%             2. Model's predictions are deducible from the explanation: num(examples where IE->O and joint have same predictions)/num(examples) \\
%     \item What kind of ablations make sense here? \\
%         1. parallel vs pipeline
%         2. with/without faithfulness
%         3. which teacher is more useful?

% \end{itemize}

\subsection{Discussion}
Next we conduct several deeper analysis of the models and provide detailed insight to the overall results presented in Section \ref{subsec:main_results}.

\subsubsection*{RQ1: Does breaking the joint into conditionals improve performance for task label prediction and explanation quality?}

%\begin{table}[tbp]
%	\centering
%	\resizebox{\linewidth}{!}{
%		\begin{tabular}{llcccc}
%			\toprule
%			& Model & e-SNLI & ComVE & ECQA & Avg \\
%			& /Metric & Acc/BLEU & Acc/BLEU & Acc/BLEU & Acc/BLEU\\
%			\midrule
%			\multirow{3}{*}{Fully Sup.}
%			& Predictor & $89.7$/$-$ & $\bf{90.2}$/$-$ & $\bf{55.9}$/$-$ & $\bf{78.6}$/$-$ \\
%			& Rationalizer & $-$/$\bf{34.9}$ & $-$/$\bf{16.8}$/ & $-$/$\bf{18.8}$ & $-$/$\bf{23.5}$ \\
%			& Joint & $\textbf{90.4}$/$33.7$ & $86.2$/$14.5$ & $53.6$/$16.2$ & $76.7$/$21.5$ \\
%			\midrule
%			\multirow{3}{*}{Few-Shot} & Predictor & $\bf{82.9}$/$-$ & $\bf{75.7}$/$-$ & $\bf{39.7}$/$-$ & $\bf{66.1}$/$-$ \\
%			& Rationalizer & $-$/$\bf{27.1}$ & $-$/$\bf{14.8}$ & $-$ /$\bf{16.3}$ & $-$/$\bf{19.4}$ \\
%			& Joint & $82.6$/$24.2$ &	$73.8$/$12.7$ &	$34.3$/$9.8$ & $63.5$/$15.6$ \\
%			\midrule
%			\multirow{3}{*}{Self-Train.} & Predictor & $\bf{83.8}$/$-$ & $\bf{78.8}$/$-$ & $\bf{44.4}$/$-$ & $\bf{69}$/$-$ \\
%			& Rationalizer & $-$/$\bf{31.2}$ & $-$/$\bf{17.0}$ & $-$/$\bf{19.2}$ & $-$/$\bf{22.5}$ \\
%			& Joint & $83.4$/$24.5$ & $79.2$/$11.0$ & $41.8$/$9.8$ & $68.1$/$15.1$ \\
%			\bottomrule
%		\end{tabular}
%	}
%	\vspace{-0.2cm}
%	\caption{Performance of the Joint model compared to Predictor and Rationalizer models in Fully supervised, Few-Shot and Self-Training setup.}
%	\label{tab:conditionals}
%	\vspace{-0.1cm}
%\end{table}

\begin{table*}[tbp]
\centering
\resizebox{0.7\textwidth}{!}{
\begin{tabular}{llcc|cc|cc|cc}
\toprule
 & Model & \multicolumn{2}{c|}{e-SNLI} & \multicolumn{2}{c|}{ComVE} & \multicolumn{2}{c|}{ECQA} & \multicolumn{2}{c}{Avg} \\
& /Metric & Acc & BLEU & Acc & BLEU & Acc & BLEU & Acc & BLEU\\
\midrule
\multirow{3}{*}{Fully Supervised}
& Predictor & $89.7$ & $-$ & $\bf{90.2}$ & $-$ & $\bf{55.9}$ & $-$ & $\bf{78.6}$ & $-$ \\
& Rationalizer & $-$ & $\bf{34.9}$ & $-$ & $\bf{16.8}$ & $-$ & $\bf{18.8}$ & $-$ & $\bf{23.5}$ \\
& Joint & $\textbf{90.4}$ & $33.7$ & $86.2$ & $14.5$ & $53.6$ & $16.2$ & $76.7$ & $21.5$ \\
\midrule
\multirow{3}{*}{Few-Shot} & Predictor & $\bf{82.9}$ & $-$ & $\bf{75.7}$ & $-$ & $\bf{39.7}$ & $-$ & $\bf{66.1}$ & $-$ \\
& Rationalizer & $-$ & $\bf{27.1}$ & $-$ & $\bf{14.8}$ & $-$ & $\bf{16.3}$ & $-$ & $\bf{19.4}$ \\
& Joint & $82.6$ & $24.2$ &	$73.8$ & $12.7$ &	$34.3$ & $9.8$ & $63.5$ & $15.6$ \\
\midrule
\multirow{3}{*}{Self-Training} & Predictor & $\bf{83.8}$ & $-$ & $\bf{78.8}$ & $-$ & $\bf{44.4}$ & $-$ & $\bf{69}$ & $-$ \\
& Rationalizer & $-$ & $\bf{31.2}$ & $-$ & $\bf{17.0}$ & $-$ & $\bf{19.2}$ & $-$ & $\bf{22.5}$ \\
& Joint & $83.4$ & $24.5$ & $79.2$ & $11.0$ & $41.8$ & $9.8$ & $68.1$ &	$15.1$ \\
\bottomrule
\end{tabular}
}
\caption{Performance of the Joint model compared to Predictor and Rationalizer models in Fully supervised, Few-Shot and Self-Training setup.}
\label{tab:conditionals}
\end{table*}

%\begin{table}[tbp]
%	\centering
%	\resizebox{\linewidth}{!}{
%		\begin{tabular}{lcccccccc}
%			\toprule
% Model & \multicolumn{2}{c|}{e-SNLI} & \multicolumn{2}{c|}{ComVE} & \multicolumn{2}{c|}{ECQA} & \multicolumn{2}{c}{Average} \\
%			 /Metric & Acc & BLEU & Acc & BLEU & Acc & BLEU & Acc & BLEU\\
%			\midrule
%			\textit{Fully-Supervised}&&&&&&&&\\ 
%			Predictor & $89.7$ & $-$ & $\bf{90.2}$ & $-$ & $\bf{55.9}$ & $-$ & $\bf{78.6}$ & $-$ \\
%			Rationalizer & $-$ & $\bf{34.9}$ & $-$ & $\bf{16.8}$ & $-$ & $\bf{18.8}$ & $-$ & $\bf{23.5}$ \\
%			Joint & $\textbf{90.4}$ & $33.7$ & $86.2$ & $14.5$ & $53.6$ & $16.2$ & $76.7$ & $21.5$ \\
%			\midrule
%			\textit{Few-Shot}&&&&&&&&\\ 
%			Predictor & $\bf{82.9}$ & $-$ & $\bf{75.7}$ & $-$ & $\bf{39.7}$ & $-$ & $\bf{66.1}$ & $-$ \\
%			 Rationalizer & $-$ & $\bf{27.1}$ & $-$ & $\bf{14.8}$ & $-$ & $\bf{16.3}$ & $-$ & $\bf{19.4}$ \\
%			Joint & $82.6$ & $24.2$ &	$73.8$ & $12.7$ &	$34.3$ & $9.8$ & $63.5$ & $15.6$ \\
%			\midrule
%			\textit{Self-Training }&&&&&&&& \\
%			Predictor & $\bf{83.8}$ & $-$ & $\bf{78.8}$ & $-$ & $\bf{44.5}$ & $-$ & $\bf{69}$ & $-$ \\
%			Rationalizer & $-$ & $\bf{31.2}$ & $-$ & $\bf{17.0}$ & $-$ & $\bf{19.2}$ & $-$ & $\bf{22.5}$ \\
%			Joint & $83.4$ & $24.5$ & $79.2$ & $11.0$ & $41.8$ & $9.8$ & $68.1$ &	$15.1$ \\
%			\bottomrule
%		\end{tabular}
%	}
%	\vspace{-0.2cm}
%	\caption{Performance of the Joint model compared to Predictor and Rationalizer models in Fully supervised, Few-Shot and Self-Training setup.}
%	\label{tab:conditionals}
%	\vspace{-0.3cm}
%\end{table}

We first want to analyze the effectiveness of breaking the joint model into conditionals and learning two separate models for task prediction and rationalization. 
From the results in Table \ref{tab:conditionals},
%shows the performance of the Joint model, in comparison to the Predictor model for task accuracy and Rationalizer model BLEU score for explanation quality.
%We report results in both few-shot and self-training settings. 
it is evident that by breaking the joint distribution into conditionals, we obtain significantly higher performance across all datasets, especially for explanation generation. This validates our hypothesis that with limited labels, it is much harder for the model to learn the joint distribution of output and explanation, compared to learning the conditionals separately. With self-training, the gap in performance between the joint and the conditionals decreases, but the individual models still outperform the joint model. 

These results align with the improvement observed from the Dual Teacher framework over Joint model in Table \ref{tab:main_results}.
Training the Predictor and Rationalizer models in their own self-training loops creates two strong teacher models and provides better pseudo labels. This allows us to train a strong self-rationalizing model through distillation than training a joint model directly through self-training.

%\textbf{write about parallel vs pipeline}

\subsubsection*{RQ2: Does the Masked Label Regularization help to generate more faithful explanations?}

% \begin{table}[htbp]
% \centering
% \begin{tabular}{l|c|c|c}
% \toprule
% Dataset & $P(O|I)$ & $P(O|I,E^*)$ & $P(O|I,E')$ \\
% \midrule
% e-SNLI & 89.73 & 98.62 & \\
% ComVE &  88.6  & 91.4 & \\
% ECQA  &  55.56 & 98.04 & \\
% \bottomrule
% \end{tabular}
% \caption{Results of different predictor models. $E^*$ are the gold explanations \textbf{Send to appendix?}}
% \label{tab:predictor_results}
% \end{table}

\begin{table}[tbp]
\centering
\resizebox{0.45\textwidth}{!}{
\begin{tabular}{l|c|c|c|c}
\toprule
Model/Dataset & e-SNLI & ComVE & ECQA & Avg\\
\midrule
% Fully-supervised & ? & ? & ? & ? \\
% Few-shot & ? & ? & ? & ? \\
%Vanilla & $76.6$ & $60.7$ & $92.8$ & $76.7$ \\
Joint & $76.8$ & $52.0$ & $89.0$ & $72.6$ \\
Dual Teacher $-$ MLR & $86.5$ & $54.8$ & $93.9$ & $78.4$ \\
Dual Teacher & $95.7$ & $74.7$ & $95.8$ & $88.7$ \\
\midrule
Rationalizer $-$ MLR & $97.8$ & $69.9$ & $95.3$ & $87.7$ \\
Rationalizer & $\mathbf{99.4}$ & $\mathbf{84.8}$ & $\mathbf{96.8}$ & $\mathbf{93.7}$ \\
\bottomrule
\end{tabular}
}
\caption{Label-Explanation association measured as \% of inputs with distinct explanations for each task label.}
\label{tab:label_variation}
\end{table}

While our method achieves better BLEU scores compared to different baselines, it is also important to evaluate whether the generated explanations are \textit{faithful} to the predictions, i.e. provide reasoning that support the predicted label. 
During creation of the datasets, the annotators were instructed to assign a label and then explain the assignments with a natural language explanation. Therefore, it is desirable for the models to preserve the faithfulness properties in generated explanations. 

We perform two tests to analyze whether (1) the explanations are dependent on the output and (2) if they reflect the intended label. Through these experiments we also conduct an ablation study to estimate the effect of the proposed Masked Label Regularization (MLR) constraint in improving the faithfulness of explanations.

\textbf{Label-Explanation Association.}
We first conduct a simple analysis to check if the explanations are dependent on the model predictions. 
As a necessary condition for generating faithful explanations, different predicted labels have to produce different explanations. We measure this association as the number of test instances for which the model generates a distinct explanation for all labels. 

We vary the task label and ask the model to generate an explanation. 
For joint models, 
%the label is generated as the first part of the output sequence. 
we replace the generated label with other possible labels and ask the decoder to continue generating an explanation.
For Rationalizer model, we simply generate predictions with providing different labels in the input. We study the effect of MLR  by removing the entropy regularization loss while training the Rationalizer. We denote this variant as Rationalizer $-$ MLR.  Dual teacher $-$ MLR refers to the Joint model trained using Rationalizer $-$ MLR.

Results in Table \ref{tab:label_variation} show that for the Joint model,  only $72\%$ of the examples have unique explanations per output on an average across datasets. This implies that the label-explanation association is not inherently captured in the decoder and for $28\%$ of instances the generated explanation is constant and has no association with the labels.
%The Dual Teacher - MLR model improves over the Joint model by a small margin. 
Adding the MLR loss encourages the model to condition on labels, and thereby provides a substantial improvement of over $10\%$ for the Dual Teacher model. This indicates a strong association between the generated label and explanation, where the explanations are unique to the label in over $88\%$ of cases.
As can be seen from the Table, the Rationalizer teacher achieves significantly better label-explanation association compared to the Joint counterparts. The MLR constraint further improves the results, especially in the ComVE dataset where explanations are much longer on average.

%\subsubsection*{Simulatibility of Explanations}

% Next, we utilize prior work and compute the simulatability of the explanations following prior work \cite{hase2020leakage, Wiedgreffe2020MeasuringAB}. We can observe that the explanations generated by our method improves shows substantial improvement in the case of e-SNLI and ECQA, and remains similar in the case of ComVE across different models. Upon examining the cases where the model performs incorrectly, we see that the model predicts the wrong label for such examples, subsequently generating an explanation which is untrue. \attn{review this}.

\begin{table}[tbp]
\centering
\resizebox{0.9\linewidth}{!}{
\begin{tabular}{l|cccc}
\toprule
Model/Dataset & e-SNLI & ECQA & ComVE & Avg\\
\midrule
Fully-Supervised & 8.64 & 30.45 & 2.4 & 13.83 \\
Few-Shot & 2.61 & 16.91 & 0.2 & 6.57 \\
\midrule
%Vanilla & 4.75 & 12.58 & 0.3 & 5.88 \\
Joint & 1.87 & 14.22 & 0.7 & 5.6 \\
Dual Teacher - MLR & \textbf{6.01} & 17.96 & 0.73 & 7.62 \\
Dual Teacher & 4.54 & \textbf{18.19} & \textbf{0.9} & \textbf{7.88} \\
\bottomrule
\end{tabular}
}
\caption{Simulatability score of the explanations from different methods. The higher the score the more aligned the explanation is with the predicted label}
\label{tab:simulatability_results}
\end{table}

\begin{figure}[tb]
\begin{subfigure}{\linewidth}
  \includegraphics[width=\linewidth]{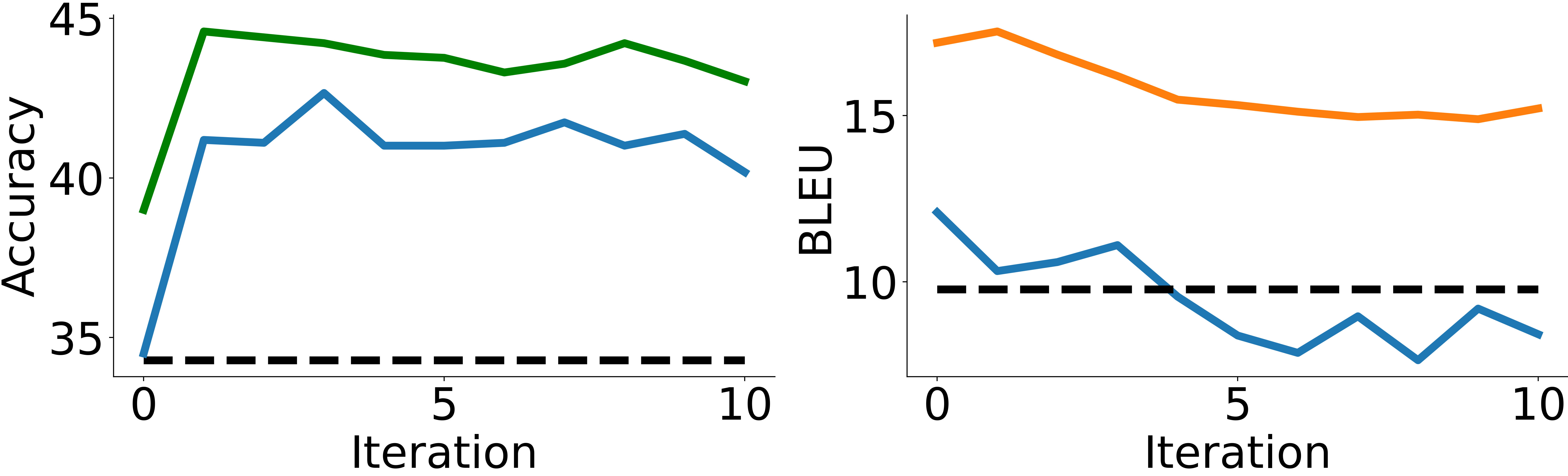}
  \caption{ECQA}
\end{subfigure}
\vfill
\begin{subfigure}{\linewidth}
  \includegraphics[width=\linewidth]{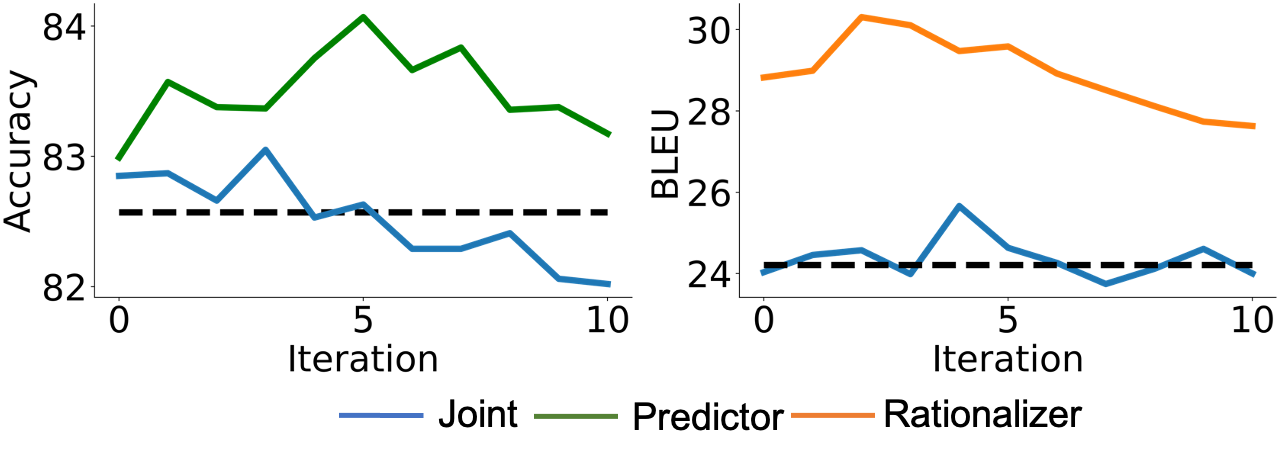}  
  \caption{e-SNLI}
\end{subfigure}
\caption{Performance across self-training iterations on ECQA and e-SNLI datasets of the Confidence Weighted Joint, Predictor and Rationalizer models. Dashed lines show the performance of the Few-Shot Joint model.}
\label{fig:self_training_progression}
\end{figure}

\begin{table*}[tb]
\centering
\small
\begin{tabular}{lcccccc}
\toprule
& \multicolumn{2}{c}{ECQA} & \multicolumn{2}{c}{ComVE} & \multicolumn{2}{c}{eSNLI} \\
Size of $D_l$ & Accuracy & BLEU & Accuracy & BLEU & Accuracy & BLEU \\
\midrule
10 & 30.63 & 16.36 & 49.6 & 15.55 & 81.97 & 24.62 \\
100 & 44.26 & 17.12 & 79.61 & 14.83 & 83.95 & 30.17 \\
200 & 47.81 & 17.01 & 80.1 & 14.45 & 83.61 & 29.59 \\
500 & 51.28 & 19.03 & 82.4 & 15.88 & 84.43 & 29.3 \\
5000 & - & - & - & - & 85.85 & 29.37 \\
\midrule
Fully supervised & 53.6 & 16.25 & 86.2 & 14.53 & 90.44 & 33.76 \\
\bottomrule
\end{tabular}
\caption{Effect of size of the labeled set on the final performance.}
\label{tab:full_label_set_size_vs_perf}
\end{table*}

\textbf{Simulatibility of Explanations.}
We utilize the Simulatability metric as defined in prior work \cite{chan2022frame, hase2020evaluating} to evaluate how well an external system, human or AI, is able to simulate the prediction made by a black-box, self-rationalizing model using the explanation it generates. 
As simulators, two models are trained to predict the task label - (1) a control model $P(O|I)$, which predicts the output given input and (2) a treatment model,  $P(O|I,E)$ predicting output given input and an explanation. 
The simulators are used to measure how much the explanations generated by the self-rationalizing model help in `guessing' its predicted label. 
The simulatability score is defined as 
\begin{equation}
    \Phi = \mathbbm{1}(y^T = \hat{y}) - \mathbbm{1} (y^C = \hat{y})
\end{equation}
\noindent where $\hat{y}$ refers to predicted label from the self-rationalizing model, $y^C$ and $y^T$ refers to predictions from the control and treatment simulators, respectively. 
%The treatment simulator makes use of explanation in order to predict the output, and therefore is expected to produce an output that aligns with the explanation. 
%During the simulatibility test, it is fed with the explanation produced from a self-rationalizing model. 
The higher the faithfulness of a model, the better aligned its explanations are with its predicted labels, relative to the control simulator which does not consider explanations.

Table \ref{tab:simulatability_results} shows the simulatibility scores of the various self-rationalizing models under consideration. We observe a similar trend as in Table \ref{tab:label_variation} while comparing the different models, with the exception of e-SNLI. 
For e-SNLI the control simulator was notably stronger compared to treatment, potentially due to the overlap with pre-training tasks of T5.
%On e-SNLI, surprisingly the self-training produced a notably worse Rationalizer, which also impacted the Dual-Teacher model's results as it relied on confidence weights.
We note that overall there is significant gap in the simulatability between our models and the Fully-Supervised model, indicating a large room for improvement in the faithfulness of explanation for weakly supervised models. 

\subsubsection*{RQ3: How does the performance change as self-training progresses?}

% \begin{figure*}[htb]
% \begin{subfigure}{0.32\linewidth}
%   \includegraphics[width=\linewidth]{figs/self-training-progression/ECQA-predictor-scores-over-self-training.png}
% \end{subfigure}

% \begin{subfigure}{0.32\linewidth}
%   \includegraphics[width=\linewidth]{figs/self-training-progression/ECQA-predictor-scores-over-self-training.png}
% \end{subfigure}

% \begin{subfigure}{0.32\linewidth}
%   \includegraphics[width=\linewidth]{figs/self-training-progression/ECQA-predictor-scores-over-self-training.png}
% \end{subfigure}

% \begin{subfigure}{0.32\linewidth}
%   \includegraphics[width=\linewidth]{figs/self-training-progression/ECQA-rationalizer-scores-over-self-training.png}
%   \caption{ComVE}\label{fig:awesome_image1}
% \end{subfigure}

% \begin{subfigure}{0.32\linewidth}
%   \includegraphics[width=\linewidth]{figs/self-training-progression/ECQA-rationalizer-scores-over-self-training.png}
% \end{subfigure}

% \begin{subfigure}{0.32\linewidth}
%   \includegraphics[width=\linewidth]{figs/self-training-progression/ECQA-rationalizer-scores-over-self-training.png}  
% \end{subfigure}
% \caption{Performance curve during self-training iterations on the three datasets}
% \end{figure*}

Figure \ref{fig:self_training_progression} shows the performance of different models over self-training iterations. We observe that the two teacher models consistently outperform the joint model over iterations in both datasets. In ECQA dataset there is a large jump in accuracy in the first iteration and the algorithm converges soon. A similar trend is observed for BLEU scores, with a slight improvement in the Rationalizer in first iteration and the score plateauing or even declining in case of the Joint model. 
For e-SNLI dataset, accuracy continues to improve till five iterations for the Predictor, and three for the Joint model. The rationalization performance also converges after nearly five iterations for both the models. 
Convergence of the algorithm could be explained by the poor separability of the class labels in the datasets, causing more erroneous pseudolabels and plateauing of performance as time progresses.

\subsubsection*{RQ4: How does the performance change with increase in labelled dataset size?}

% \begin{table}[tb]
% \centering
% \small
% \begin{tabular}{l|cc}
% \toprule
% Size of $D_l$ & Accuracy & BLEU \\
% \midrule
% 50 & 30.63 & 16.36 \\
% 500 & 44.26 & 17.12 \\
% 1000 & 47.81 & 17.01 \\
% 2500 & 51.28 & 19.02 \\
% \midrule
% Fully supervised & 53.6 & 16.25 \\
% \bottomrule
% \end{tabular}
% \vspace{-0.2cm}
% \caption{Effect of size of the labeled set on the final performance on the ECQA dataset.}
% \label{tab:label_set_size_vs_perf}
% \vspace{-0.3cm}
% \end{table}

We study the performance of our model by conducting experiments with different dataset sizes. We only vary the labeled dataset size and keep the remaining training set as unlabeled data. For example, for ECQA the total size ($D$) is $7.5K$, and we conduct experiments with labeled data ($D_l$) in the range $\{50, 2.5K\}$ and the remaining data size ($D-D_l$) as unlabeled data.
Table \ref{tab:full_label_set_size_vs_perf} reports the accuracy and BLEU score of our proposed model for dataset sizes ranging from $50$ to $2500$ samples. 
We see that there is a improvement in the test accuracy and BLEU score as the labeled data size increases. 
% The model is able to achieve an accuracy within 2\% of the fully supervised model using only one-third of the data. 
With as few as 500 examples per label, the model is able to achieve accuracy within 6\% of the fully supervised model across all datasets.
%Similarly for BLEU score we observe that the performance improves with larger labeled dataset. 
Interestingly, we note that with limited supervision the self-training setup is able to outperform the Fully supervised model in the BLEU scores, demonstrating the data efficiency of the Rationalizer teacher in achieving good performance.

\section{Conclusion}
We study the self-rationalization problem with few-shot labels and demonstrate that self-training is an effective learning paradigm and can significantly reduce the gap between few shot and fully supervised model performance. 
We present a novel dual teacher learning framework that learns two models for task label prediction and rationale generation through self-training and efficiently distill the knowledge in a single self-rationalizing joint model.
With a masking based loss formulation we enforce label-explanation association in the rationalizer, leading to generation of more faithful explanations.
We conduct experiments on three public benchmark datasets for free text explanations, and show that the proposed methods are effective in improving task performance while generating accurate and faithful explanations.

\section{Limitations}
Despite strong performance compared to few-shot our self-training methods still contain significant room for improvement compared to the fully supervised benchmarks. It would be interesting to try larger language models to see if it is possible to close this gap with more knowledge embedded into the pre-trained models.
Our evaluation of free text rationales are limited by the automatic metrics, which are necessary but not sufficient to analyze quality of an explanation for decision making of the model. From example explanations (a few of which are shown in Appendix), it is evident that we still lack understanding on multiple dimensions such as, when an explanation is factually wrong, is it due to the model believing in the wrong knowledge or is unable to retrieve the correct one. Works that probe a language model with various prompts could be useful for investigating in these directions.

\bibliography{references}
\bibliographystyle{acl_natbib}

\section{Appendix}
% \textbf{Add a case-study section}
% predictor got it wrong -> impact on rationalizer -> impact on joint
% breaking into conditionals: wrong prediction -> wrong explanation
%                             correct -> correct explanation 

% Better inductive bias through the Rationalizer by moving  the label to the input
% take the correct predictions, pass them through the rationalizer get a few examples

% Sensitivity to input labels:

% label variation example in the Joint -- comparative

% 1. moving the prediction to the encoder (Rationalizer) is better inductive bias
%     Rationalizer > Joint
% 2. Faithfulness regularization
%     label variation example 
%         Joint (one example from e-SNLI)
% 3. Model comparison
%     Fundamental problems
%         wrong commonsense by rationalizer
%         the predictor making a mistake and the rationalizer explained the wrong label in a nonsensical manner.

We include some qualitative analysis of the different design choices in our method.
% namely the introduction of the Rationalizer teacher model, which we use to shift the label to the encoder and helping the T5 model to encode it better. In addition, we demonstrate the effect of the Masked Label Regularization in inducing the dependency on the task label. We conclude this section with an error analysis highlighting the limitations of the current setup.

\begin{table*}[tb]
\centering
\resizebox{0.9\textwidth}{!}{
\begin{tabular}{l}
\toprule
\textbf{Input}: \textit{premise}: a woman plays guitar for tips. \textit{hypothesis}: lessons are there to learn  \\
\textbf{Label}: neutral  \\
\textbf{Gold}: a woman playing a guitar does not imply she is giving lessons. \\
\textbf{Joint}: a woman can be a teacher and not be a teacher. \\
\textbf{Rationalizer}: just because a woman plays guitar does not mean they are there to learn. \\
\midrule
\textbf{Input}: \textit{choice1}: she turned in her homework before it was due and got full credit \textit{choice2}: she turned in her homework five years after it was due and got full credit \\
\textbf{Label}: choice2 \\
\textbf{Gold}: homework must typically be turned in before the due date in order to receive full credit \\
\textbf{Joint}: homework is not done when it is due. \\
\textbf{Rationalizer}: you cannot turn in your homework five years after it was due \\
\midrule
\textbf{Input}: what will friends in elementary school likely do? \textit{choice}: go somewhere \textit{choice}: tell story \textit{choice}: going out at night \textit{choice}: jump rope \textit{choice}: teach \\
\textbf{Label}: jump rope \\
\textbf{Gold}: kids in elementary school generally play games like hide and seek, tag, musical chairs and so on. \\
\textbf{Joint}: jump rope is a rope used to teach others to jump rope. friends in elementary school likely will do something like jumping rope. \\
\textbf{Rationalizer}: friends in elementary school will jump rope. \\
\bottomrule
\end{tabular}}
\caption{Sample explanations generated from a few-shot Joint model as compared to a few shot Rationalizer model.}
\label{tab:rationalizer_vs_joint_explanations}
\end{table*}

\begin{table*}[tb]
	\centering
	\resizebox{0.7\linewidth}{!}{
		\begin{tabular}{l}
			\toprule
			\textit{premise}: two little boys wearing athletic jerseys are washing their hands in a public restroom. \\
			\textit{hypothesis}: they were playing soccer. \\
			\textbf{With MLR}\\
			\textbf{entailment} explanation: boys are washing their hands while playing soccer. \\
			\textbf{neutral} explanation: just because boys are wearing athletic jerseys, it does not mean they are playing soccer.\\
			\textbf{contradiction} explanation: boys cannot be wearing athletic jerseys and playing soccer at the same time. \\
			\midrule
			\textbf{Without MLR} \\
			\textbf{entailment} explanation: boys are boys.\\
			\textbf{neutral} explanation: boys cannot be washing their hands while playing soccer. \\
			\textbf{contradiction} explanation: boys cannot be washing their hands while playing soccer. \\
			
			\midrule
			\midrule
			\textit{choice1}: bats can fly perfectly. \textit{choice2}: bats can ride bicycles. \\
			
			\textbf{With MLR} \\
			\textbf{choice1} explanation: bats cannot fly. \\
			\textbf{choice2} explanation: bats cannot ride bicycles. \\
			\midrule
			\textbf{Without MLR} \\
			\textbf{choice1} explanation: bats cannot ride bicycles. \\
			\textbf{choice2} explanation: bats cannot ride bicycles. \\
			
			\bottomrule
	\end{tabular}}
	\caption{Case studies of generated explanations for varied task labels with and without the MLR loss constraint}
	\label{tab:effect_of_mlr_regularizer}
\end{table*}

\begin{table*}[tb]
	\centering
	\resizebox{0.8\textwidth}{!}{
		\begin{tabular}{l}
			\toprule
			% \textbf{Input}: \textit{premise}: a young girl is playing a musical instrument and singing into a microphone. \textit{hypothesis}: a little girl belts one out while playing the violin. \\
			% \textbf{Label}: neutral \\
			% \textbf{Gold Explanation}: just because she is playing a musical instrument, it doesn't mean that instrument is a violin. \\
			% \textbf{Prediction}: contradiction explanation: a young girl cannot be playing a musical instrument and singing into a microphone at the same time. \\
			
			% \midrule
			\textbf{Input}: choice1: she shaved her eyes. choice2: she shaved her legs. \\
			\textbf{Label}: choice1  \\
			\textbf{Gold Explanation}: there is no hairs to shave on the eyes. \\
			\textbf{Predicted Label} choice2 \\
			\textbf{Generated explanation:} legs are not razors. \\
			\midrule
			
			\textbf{Input}: cats have how many appendages? choice: tail choice: whiskers choice: two eyes choice: four paws choice: four legs \\
			\textbf{Label} four legs \\
			\textbf{Gold Explanation}: appendage refers to something that is attached four legs are attached to cats four legs are used to walk \\
			\textbf{Predicted Label} four paws \\
			\textbf{Generated explanation}: four paws are appendages. cats have two eyes. \\
			
			\bottomrule
	\end{tabular}}
	\caption{Qualitative analysis of the prediction errors of our model.}
	\label{tab:failure_cases}
\end{table*}

\subsection{Impact of moving the task label to the input from the output sequence}
We observed substantial improvement in rationale performance with the Rationalizer teacher model compared to the Joint model. 
This is can be attributed to the the prediction being passed as an input to encoder of the Seq2Seq model, generating better representation of the predictions and yielding better quality rationales. 
%This is even more important in the case of few-shot settings, as the better quality pseudolabels are needed for obtaining an improvement in the performance. 
Table \ref{tab:rationalizer_vs_joint_explanations} shows a sample of explanations generated from the Joint and Rationalizer models, for cases when the label was predicted correctly.
We see that the Rationalizer generally produces higher quality explanations, and in contrast, while the Joint model often generates nonsensical explanations with frequently repeated words. 
The better quality rationales obtained from the Rationalizer teacher helps generate better pseudo labels and the final model is able to capture those with distillation. 

\subsection{Effect of the Masked Label Regularization on faithfulness}

Table \ref{tab:effect_of_mlr_regularizer} shows a few examples of rationales generated by the Dual Teacher model with and without the MLR loss. From the first example on e-SNLI dataset, we see that without MLR constraint the model generates same explanation for neutral and contradiction labels, and the explanation for the neutral label indicates a contradiction. In contrast when trained with MLR, it outputs an explanation which is in alignment with the assigned label. 
In the second example from ComVE, the model without MLR outputs the same explanation showing that it ignores the label assigned. 
With MLR constraint the model is able to generate explanations sensitive to the assigned label. Although the reasoning for the incorrect label is wrong, this behavior is still desired for an interpretable system elucidating  why a prediction was made.

\subsection{Error Analysis}

Table \ref{tab:failure_cases} shows a snapshot of the qualitative analysis of the errors from our model. From the explanations generated for the predictions, we see that the model is unaware of situations which require additional background information, such as the existence of hair on eyes, or subtle differences between words, such as paws and feet. We believe a better pretrained Language Model can help alleviate some of these issues.

\end{document}